\documentclass[11pt]{article}

\usepackage[final]{acl}

\usepackage{times}
\usepackage{latexsym}

\usepackage[T1]{fontenc}

\usepackage[utf8]{inputenc}

\usepackage{microtype}

\usepackage{inconsolata}

\usepackage{graphicx}

\usepackage{booktabs}
\usepackage{multirow}
\usepackage{adjustbox}
\usepackage{amsmath}
\usepackage{xcolor}
\usepackage{soul}
\sethlcolor{gray!15} 
\usepackage[most]{tcolorbox}
\usepackage{caption} 
\usepackage{capt-of}
\usepackage{needspace}

\usepackage{silence}
\title{Beyond Length: Context-Aware Expansion and Independence \\as Developmentally Sensitive Evaluation in Child Utterances}

\author{
 \textbf{Jiyun Chun},
 \textbf{Eric Fosler-Lussier},
 \textbf{Michael White},
 \textbf{Andrew Perrault}
\\
The Ohio State University\\
\texttt{\{chun.203,fosler-lussier.1,white.1240,perrault.17\}@osu.edu}\\
\\
}

\newcommand{\Exp}{\textit{E}}
\newcommand{\Ind}{\textit{I}}
\newcommand{\PT}{\textit{PT}}

\begin{document}
\nolinenumbers
\maketitle
\begin{abstract}
Evaluating the quality of children's utterances in adult-child dialogue remains challenging due to insufficient context-sensitive metrics. Common proxies such as Mean Length of Utterance (MLU), lexical diversity (vocd-D), and readability indices (Flesch-Kincaid Grade Level, Gunning Fog Index) are dominated by length and ignore conversational context, missing aspects of response quality such as reasoning depth, topic maintenance, and discourse planning. 
We introduce an LLM-as-a-judge framework that first classifies the \textbf{Previous Adult Utterance Type} and then scores the child's response along two axes: \textbf{Expansion} (contextual elaboration and inferential depth) and \textbf{Independence} (the child's contribution to advancing the discourse). 
These axes reflect fundamental dimensions in child language development, where Expansion captures elaboration, clause combining, and causal and contrastive connectives. Independence captures initiative, topic control, decreasing reliance on adult scaffolding through growing self-regulation, and audience design.
We establish developmental validity by showing age-related patterns and 
demonstrate predictive value by improving age estimation over common baselines. 
We further confirm semantic sensitivity by detecting differences tied to 
discourse relations. Our metrics align with human judgments, enabling large-scale evaluation. This shifts child utterance assessment from simply measuring length to evaluating how meaningfully the child's speech contributes to and advances the conversation within its context.\footnote{\url{https://github.com/Amyyyyeah/Beyond-Length}}


\end{abstract}

\section{Introduction}


Child-facing AI systems are increasingly deployed in settings such as educational chatbots, museum guides, and language learning tools \citep{vociechatbot, aichat, museumai, languagelearning}. In these applications, designers and researchers need ways to evaluate child utterances beyond surface length, capturing how a child's turn fits the local question-answer relation and how it advances discourse. 
Beyond age appropriateness, our broader aim is to develop context-aware metrics that capture developmentally relevant signals in real child-adult conversations.

Evaluating children's language development in dialogue remains challenging. Widely used proxies such as Mean Length of Utterance (MLU; \citealt{MLU}), type-token ratios (TTR; \citealt{ttr}), and readability formulas including Flesch-Kincaid Grade Level \citep{FKGL} and Gunning Fog Index \citep{GF} were designed for written text rather than spontaneous child-adult conversation. Length-based measures capture quantity but often fail to distinguish long yet content-poor utterances from brief yet well-reasoned ones. TTR variants are sensitive to sample size \citep{LimitTTR, limitTTR2}. vocd-D reduces but does not remove this sensitivity, remaining unreliable on very short turns \citep{LimitVOCD,LimitLexical}. Readability indices depend on sentence boundaries and syllable counts, which makes them brittle in the presence of disfluencies, fragments, and backchannels that are common in child speech \citep{LimitRead, LimitRead2, childes, disfluency}. More broadly, these proxies tend to be context insensitive because they ignore how a child's turn relates to the preceding adult prompt such as question type and discourse function. They also correlate strongly with length, which encourages a length as proxy interpretation where verbosity is mistaken for competence. We therefore need evaluation measures that are sensitive to semantic and pragmatic contributions in context and that can disentangle substantive development from mere increases in utterance length.

A more fundamental limitation is the lack of context awareness. In child-adult dialogue, a child's response is tightly linked to the immediately preceding adult turn (contingency; \citealt{BLOOM1976521, Rowe2012ALI, callanan2020exploration, LDR511, teacher}). Two responses of similar length may differ greatly in inferential depth and alignment to the question. For example, to "What did you do at school today?", "I don't really remember. We just hung out at recess, nothing special, I guess" is vague and weakly aligned, whereas "We started a bean-germination experiment, labeled our pots, and we'll check for sprouting." conveys concrete actions, temporal structure, and a plan, thereby expanding the prior context. Metrics that ignore question--answer relations and discourse functions (e.g., topic maintenance, repair, joint planning) tend to underestimate such developmental signals \citep{wh, Clark02072020, develp}.

Psycholinguistics and developmental linguistics have established concepts of conversational development: contingency (contextually appropriate responses), elaboration (addition of details and reasoning), scaffolding (utterance facilitation), and topic maintenance \citep{BLOOM1976521, fivush2006elaborating, ninio1996pragmatic}. These frameworks  can be summarized along two axes: children's ability to independently advance discourse and their capacity to expand upon previous turns' meaning \citep{SALOMO_LIEVEN_TOMASELLO_2013, LUO2022109}. Nevertheless, quantitative metrics for automatically measuring these concepts at scale have been absent. Prior work mainly relied on rule-based pipelines \citep{childes, miller2012assessing} or supervised classifiers \citep{sp,stolcke-etal-2000-dialogue}. Rule-based pipelines were brittle to transcription artifacts such as disfluencies and incomplete clauses and transferred poorly across datasets or recording conventions \citep{britte, MLU_2}. Supervised classifiers captured local surface features but struggled to condition on the full question-answer relation and discourse functions without heavy feature engineering, and they required retraining whenever the evaluation rubric changed. 
Recent advances in large language models (LLMs) change this feasibility boundary. LLMs can follow natural language rubrics, read context, and map free text to interpretable ordinal ratings without task-specific fine-tuning \citep{zheng2023judgingllmasajudgemtbenchchatbot, gu2025surveyllmasajudge, liu-etal-2023-g}.

We adopt an LLM-as-a-judge framework that classifies adult prompts by \textbf{Previous Adult Utterance Type} (\PT: question and non-question subtypes) and evaluates the corresponding child responses on two theoretically-motivated dimensions: \textbf{Expansion} (contextual elaboration and inferential depth) and \textbf{Independence} (autonomous contribution to discourse). To our knowledge, this study is the first to enable large-scale, context-conditioned scoring of child utterances and to demonstrate both developmental signal and predictive utility while controlling for utterance length and \PT. Empirically, our scores show systematic variation across \PT\ and remain significantly associated with age even after controlling for length and \PT. This overcomes limitations of length-dominated proxies, which fail to distinguish long-but-thin from short-but-well-reasoned responses. Moreover, at fixed utterance length, our metrics capture differences in reasoning depth and contextual fit, shifting evaluation from quantity to quality of discourse contribution. We further show that our approach provides effective predictive features for age and exhibits strong agreement with human evaluation, offering rubric alignment and interpretability that length/syllable-based metrics lack. All analyses in this paper are text-based, evaluating orthographic transcripts rather than audio. Incorporating acoustic and prosodic features is left for future work.

\section{Related Work}
Child language research has largely relied on large-scale transcript corpora such as CHILDES \citep{childes} and on standard sentence- or token-level metrics. Representative measures include Mean Length of Utterance (MLU; \citealt{MLU_1, MLU_2, MLU_3}), lexical diversity (TTR; \citealt{ttr_1, ttr_2}, vocd-D; \citealt{vocd_1, vocd_2}), and readability indices (Flesch-Kincaid; \citealt{FKGL_1, FKGL_2}, Gunning Fog; \citealt{GF_1}), with some studies supplementing clause- or phrase-based complexity indicators \citep{clauses}. However, these metrics suffer from three key limitations. First, length and complexity are confounded, making interpretation difficult. Second, stable estimation requires lengthy texts. Third, they fail to directly account for discourse context, such as the type of preceding adult question. We propose discourse-level metrics that separately measure content Expansion (\Exp) and Independence (\Ind) in child utterances.
These definitions align with established strands of child development research: \Exp\ parallels discourse elaboration \citep{EVERSVERMEUL20111645, veen, haliday, Interaction, McCabe1996RelatingEI, clauses, Kallay} in usage-based and narrative work (clause density, connective use, and detail), whereas \Ind\ tracks children's development from adult-dependent responses to self-initiated contributions, a developmental trajectory predicted by scaffolding and topic-management research \citep{Interaction, Inde,VANDAM2022101338, MASEK2021100961, motherinteraction, turntaking, children9091407, leech2013father, Chouinard2007ChildrensQA}.

Child language development extends beyond mere vocabulary accumulation to encompass the ability to expand and organize information within discourse and to advance conversation proactively. Adult interaction patterns, question types, and feedback strategies exert strong influence on the form and function of child utterances \citep{Interaction, Rowe2012ALI, callanan2020exploration, Rowe2019AnalyzingIQ, SALOMO_LIEVEN_TOMASELLO_2013}. For instance, yes/no and display questions (where asker already knows the answer) tend to elicit brief responses \citep{Display}, whereas referential and open questions, along with utterances incorporating contrast or causal markers and initiative-eliciting moves, promote longer explanations and the introduction of new information \citep{museum, raisingchildren_language_0_8}. 
Although existing research has qualitatively noted that adult utterances shape children's immediate responses, no prior work has simultaneously: (1)~quantified contextual effects at the utterance level, (2)~disentangled developmental signals from length, and (3)~demonstrated that these signals remain valid in predictive tasks such as age estimation.

\section{Methods}
\paragraph{Datasets}
We used corpora from the Child Language Data Exchange System (CHILDES \citep{macwhinney2014childes}), a large global archive of manually transcribed child--caregiver conversations ranging from birth to age 13. We focused on the English North American (Eng-NA) subcorpus and excluded sessions involving clinical populations and non-spontaneous interactions such as singing, verse recitation, and alphabet or number reading. 
As detailed in Appendix~\ref{app:dataset}, our study aggregates data from 20 distinct corpora (e.g., Brown, Clark) within CHILDES, comprising over 360K utterances and exhibiting high variance in speakers and recording conditions.
Our target age range is 2--10 years, as this period captures the most active phase of syntactic and pragmatic development.


\paragraph{Previous Adult Utterance Type}
Research on child language shows that children's turns are strongly conditioned by the discourse function of the preceding adult utterance, which can be a question, a non-question, or a special case.
We standardize the Previous Adult Utterance Type (\PT) to capture how the pragmatic demand of the immediately preceding adult move shapes children's responses. Caregivers use different kinds of questions to scaffold participation, narrowing or widening the response space. 
Closed questions (e.g., yes/no, repetition) elicit brief forms, while open questions invite elaboration and child-initiated contributions.
As shown in Table~\ref{tab:exp-indep-summary}, we classify adult utterances into questions and non-questions, each with further subdivisions. We use \verb|Gemini 2.5 Pro| \citep{comanici2025gemini25pushingfrontier} for automated annotation based on these \PT\ types. 
Detailed definitions and supporting literature for each \PT\ category are provided in Appendix~\ref{app:pt}.



\paragraph{Expansion}
To measure developmental sophistication beyond mere length, Expansion (\Exp) quantifies contextual elaboration and inferential depth, that is, how narratively and logically organized a child utterance is, with scores from 1 to 10. This definition aligns with established strands of discourse 
elaboration research examining clause density, connective 
use, and narrative detail \citep{EVERSVERMEUL20111645, veen, haliday, Interaction, McCabe1996RelatingEI, clauses, Kallay}. 
Unlike length-based proxies that fail to distinguish long-but-thin from short-but-well-reasoned responses, \Exp\ rewards logical organization and discourse linkage. In brief, the core scale progresses from single word (1), single clause (4), coordinating linkage (6), single causal relation (7), temporally ordered sequence (8), complex structure with subordination (9), to integrated narrative across multiple times or perspectives (10). 
Judgments rely on coordinating or subordinating connections, on explicit markers of cause, condition, or contrast, and on temporal or logical ordering.\footnote{Detailed scoring criteria and rubric definitions for Expansion and Independence are provided in Appendix~\ref{app:EI}. \label{fn:labeling}}

\paragraph{Independence}
Beyond responding appropriately, children gradually learn to manage and direct conversation independently. Independence (\Ind) measures the extent to which the child advances and regulates the interaction without relying on the adult context with scores from 1 to 10. \Ind\ quantifies children's progression toward autonomous conversational control, aligning with findings on scaffolding reduction and self-directed topic management \citep{Interaction, Inde,VANDAM2022101338, MASEK2021100961, motherinteraction, turntaking, children9091407, leech2013father, Chouinard2007ChildrensQA}. 
Providing a correct answer alone does not increase independence. We prioritize interactional functions such as unsolicited justification, self-initiated exploratory questions, topic management, and listener-oriented regulation. Scores progress from echoing and minimal response (1-4), through self-initiated questions and justifications (5-7), to multi-act regulation and conversational leadership (8-10).\footref{fn:labeling}

\section{Results and Analysis}
We first examine the distribution of these dimensions and demonstrate that they capture distinct developmental mechanisms rather than merely tracking utterance length. We then assess their developmental validity through an age prediction task, where we also compare against open-weight models. Additionally, we show that our dimensions can differentiate contextual nuances even when utterance lengths are similar. Finally, we validate that our automated annotations align with human judgments via human evaluation (Section~\ref{app:humeval}).

\begin{table*}[ht]
\small
\centering
\begin{adjustbox}{max width=\textwidth}
\begin{tabular}{l l c cccc cccc}
\toprule
\textbf{\PT} & \textbf{Type} & \multicolumn{1}{c}{} &
\multicolumn{4}{c}{\textbf{Expansion}} &
\multicolumn{4}{c}{\textbf{Independence}} \\
\cmidrule(lr){4-7} \cmidrule(lr){8-11}
 &  & \textbf{$N$} & Mean & SD & Median & IQR & Mean & SD & Median & IQR \\
\midrule
\multirow{8}{*}{Question}
 & Yes/No        & 25883 & 2.64 & 2.06 & 2 & 3 & 4.39 & 1.63 & 4 & 2 \\
 & Referential   & 16924 & 3.60 & 2.20 & 4 & 3 & 4.46 & 1.42 & 4 & 1 \\
 & Clarification Request  &  3586 & 3.53 & 2.09 & 3 & 2 & 4.76 & 1.58 & 4 & 1 \\
 & Display       & 11893 & 2.68 & 1.77 & 2 & 3 & 4.21 & 1.17 & 4 & 0 \\
 & Confirmation Check     & 11438 & 2.85 & 2.14 & 2 & 3 & 4.52 & 1.77 & 4 & 2 \\
 & Choice       &  2012 & 3.11 & 2.06 & 3 & 3 & 4.36 & 1.20 & 4 & 0 \\
 & Elaboration   &  6148 & 4.46 & 2.46 & 4 & 3 & 4.66 & 1.42 & 4 & 1 \\
\midrule
\multirow{4}{*}{Non-Question}
 & Topic Introduction     & 19189 & 3.25 & 2.02 & 3 & 3 & 4.94 & 1.91 & 5 & 2 \\
 & Imperative/Request     & 14759 & 3.18 & 2.01 & 3 & 3 & 4.81 & 1.91 & 5 & 1 \\
 & Feedback/Evaluation    & 39235 & 3.75 & 2.19 & 4 & 3 & 5.16 & 1.89 & 5 & 2 \\
 & Self-Talk/Thinking Aloud & 1815 & 3.38 & 2.05 & 3 & 2 & 5.10 & 1.93 & 5 & 2 \\
\bottomrule
\end{tabular}
\end{adjustbox}
\caption{Summary of Expansion and Independence scores by \PT\  (all annotated by Gemini 2.5 Pro). $N$ indicates the number of adult turns.}
\label{tab:exp-indep-summary}
\end{table*}

\subsection{Distributions of Expansion and Independence}
We examine how Expansion (\Exp) and Independence (\Ind) distribute across question and non-question types.
Table~\ref{tab:exp-indep-summary} presents descriptive statistics and the number of utterances $N$ for each previous adult utterance type (\PT). The reported statistics (mean, standard deviation, median, inter-quartile range) are computed over the child utterances that respond to those adult turns.
As expected, open-ended questions tend to elicit richer child elaboration. The mean of Expansion is highest for \emph{Elaboration}, while closed-question types such as \emph{Yes/No} and \emph{Display} exhibit lower means.
For Independence, non-question turns show higher central tendency than question turns. In our data, non-question types have higher means and medians (4.81-5.16; 5) than question types (4.21-4.76; 4).
Conversely, constrained question types limit opportunities for autonomous child talk. This constraint is reflected in \Ind, which shows narrow dispersion for \emph{Display} and \emph{Choice} questions (IQR=0). In contrast, open-ended \emph{Elaboration} questions show the largest spread in \Exp, reflecting greater response variability. 


\subsection{Developmental Validity}
\label{sec:dev-validity}
Do our metrics diagnose the kinds of changes theoretically expected with age, thereby establishing developmental validity? \Exp\ is intended to reflect development that is realized through lengthened and better-organized discourse (multi-clause integration, causal/temporal linking). By contrast, \Ind\ is designed to capture context-sensitive discourse management (initiative, clarification, regulation of topic flow) that can improve even when responses are not longer. Because chronological age co-varies with verbosity, naive age effects can collapse to longer utterances rather than genuine strategy changes. We therefore separate length-carried effects from length-independent ones and test whether \Exp/\Ind\ behave in line with these theoretical roles. All analyses in this section use \PT, \Exp, and \Ind\ scores labeled by \verb|Gemini 2.5 Pro|.

\paragraph{Setup} 
As baseline comparators, we include Mean Length of Utterance (MLU; \citealt{MLU}), lexical diversity (vocd-D; \citealt{vocd}), and readability indices (Flesch-Kincaid Grade Level; \citealt{FKGL}, Gunning Fog Index; \citealt{GF}). 
Since age and utterance length tend to increase together, the association between age and any outcome can collapse to a pure length effect unless length is separated. Yet length is necessary but not sufficient for development. Even at the same length, children can differ in discourse strategy (reasoning, clarification, topic management) and in their sensitivity to \PT. Length-based metrics and readability formulas conflate verbosity with strategy, lack context sensitivity, and become unstable when length distributions differ across children.



To avoid conflating verbosity with development, we decompose length into within-child and between-child components, and adopt a two-model design that separates effects carried by length from effects that remain once length is held constant.
Accordingly, we estimate a total-effect model (Equation~\ref{eq:total_effect}) to assess the aggregate age association and a direct-effect model (Equation~\ref{eq:direct_effect}) to assess what remains once length is held constant. 

Equations~\ref{eq:total_effect},\ref{eq:direct_effect} define the shared linear predictor $\eta$.
For \Exp/\Ind, we fit proportional odds cumulative link mixed models \textbf{CLMMs} and report odds ratios \(e^{\beta}\). For baselines, we use linear mixed models \textbf{LMMs} and report effects as coefficients \(\beta\).%
\footnote{We fit proportional \textbf{CLMMs} ($P(Y \le r)=\operatorname{logit}^{-1}(\theta_r-\eta)$) for \Exp/\Ind\ and \textbf{LMMs} ($Y=\eta+\varepsilon$) for baselines.\label{fn:lmm}} 


\begin{equation}
\label{eq:total_effect}
Y = \beta_0
+ \beta_{\text{age}}\,\text{Age}_z
+ \beta_{\text{PT}}^{\top}PT
+ u_{\text{child}} + \varepsilon
\end{equation}

\begin{equation}
\label{eq:direct_effect}
\begin{aligned}
Y = \beta_0
&+ \beta_{\text{age}}\,\text{Age}_z
+ \beta_{w}\,L^{(w)}_z
+ \beta_{b}\,L^{(b)}_z \\&
+ \boldsymbol{\beta}_{\text{PT}}^{\top}PT
+ u_{\text{child}} + \varepsilon
\end{aligned}
\end{equation}


We use $\beta$ to denote regression coefficients and $Y$ for the outcome variable (metric). \(\text{Age}_{z}\) is standardized age, and \PT\ encodes Previous Adult Utterance Type as fixed effects. We decompose utterance length into two \textit{z}-scored components computed within child\(\times\)\(\text{PT}\) cells. \(u_{\text{child}}\) is a child-level random intercept and \(\varepsilon\) is the residual error. 
Intuitively, the \textbf{within-child} component \(L^{(w)}_{z}\) captures how long the current utterance is relative to that same child's typical length under the same \PT\ (turn-level deviation from the child–\PT\ mean). 
It lets us test prompt-contingent, discourse-level adjustments at a fixed habitual length. 
By contrast, the \textbf{between-child} component \(L^{(b)}_{z}\) summarizes the child's habitual verbosity (the child–\PT\ mean length itself), reflecting how talkative that child is compared to peers under the same \PT. 
Separating these two components prevents verbosity from being mistaken for development.


\begin{table*}[t]
\centering
\small
\setlength{\tabcolsep}{10pt}
\begin{tabular}{lcll}
\toprule
\textbf{Baselines (LMMs)} & \textbf{Model} & \textbf{Age Effect ($\beta$)} & \textbf{Length Effect ($\beta$)} \\
\midrule
\multirow{2}{*}{Mean Length of Utterance (MLU)} 
    & Total & $\beta_{\text{age}}$ = 0.355 & -- \\
    & Direct & -- & self-prediction \\[4pt]
\multirow{2}{*}{Vocabulary Diversity–D (vocd-D)} 
                                & Total  & $\beta_{\text{age}}$ = 16.07 & -- \\
                                & Direct & $\beta_{\text{age}}$ = 17.00 & $\beta_{w}\approx 0$ (n.s.), $\beta_{b}=-3.22$ \\[4pt]
\multirow{2}{*}{Flesch-Kincaid Grade Level} 
     & Total & $\approx 0$ & -- \\
     & Direct & $\approx 0$ & $\beta_{w}\text{ n/a},\ \beta_{b}=2.24$ \\[4pt]
\multirow{2}{*}{Gunning Fog Index} 
    & Total & $\approx 0$ & -- \\
    & Direct & $\approx 0$ & $\beta_{w}\text{ n/a},\ \beta_{b}=1.12$ \\
\midrule
\textbf{Our metrics (CLMMs)} & \textbf{Model} & \textbf{Age Effect (OR)} & \textbf{Length Effect (OR)} \\
\midrule
\multirow{2}{*}{Expansion} & Total & OR = 1.28 ($\beta_{\text{age}}$ = 0.248) & -- \\
              & Direct & OR = 0.99 (n.s.) & $\text{OR}_{w}$ = 10.36, $\text{OR}_{b}$ = 4.27 \\[4pt]
\multirow{2}{*}{Independence} & Total & OR = 1.05 ($\beta_{\text{age}}$ = 0.053) & -- \\
                 & Direct & OR = 0.90 ($\beta_{\text{age}}$ = -0.107) & $\text{OR}_{w}$ = 2.71, $\text{OR}_{b}$ = 1.76 \\
\bottomrule
\end{tabular}
\caption{Baselines (LMM) vs.\ our metrics (CLMM) under Total vs.\ Direct specifications.  Analyses use \PT, \Exp, and \Ind\ labeled by Gemini 2.5 Pro. LMM reports $\beta$, CLMM reports odds ratios (OR). 
n.s.\ $=$ $p{>}0.05$ (not significant).}
\label{tab:trad-vs-ours}
\end{table*}

\paragraph{Analysis}
 MLU indicates typical utterance length, defined as the average number of words (or morphemes) per utterance computed over multiple turns.
In Table~\ref{tab:trad-vs-ours}, children produce longer utterances as they get older, so in the total-effect model, MLU increases with age with a small slope ($\beta$=0.355). When we include length in the model to separate the influence of length itself, the age effect disappears and only self-prediction remains. Because MLU is length by definition, it captures the pattern that speech becomes longer with age but it does not isolate differences in strategy or context beyond length.
The lexical diversity measure vocd-D appears to increase strongly with age at first glance (Total=16.07, Direct=17.00). However, the interpretation changes once we split length in two ways. vocd-D responds little to momentary within-child changes in length relative to a child's own mean ($\approx0$), while it responds negatively to differences in average length between children (-3.22). This means that children who are on average more verbose tend to receive lower diversity estimates, so when length distributions differ across groups the age-diversity relationship can be distorted. In short, vocd-D is insensitive to whether a child happens to speak longer in a given moment and tends to be biased depending on how long that child speaks on average. 
The readability indices FKGL and GFI show little response to age. Instead, scores increase for children with longer average utterances (FKGL=2.24, GFI=1.12). These indices therefore act mainly as proxies for average length rather than reflecting strategy or context. 
Expansion is intended to capture development that is realized through length. Consistent with this intention, the total-effect model shows that Expansion increases with age (1.28). In the direct-effect model where length is separated, the age effect is not significant (0.99), while the length components are very large. A child's score rises sharply when they speak longer than usual on a given turn (10.36), and children who are on average more verbose also tend to score higher (4.27). In other words, increases in Expansion are realized almost entirely through becoming longer, and children who are on average more verbose also tend to obtain higher \Exp\ scores. 
Independence is designed to capture context and strategy beyond length. Although scores increase with length (2.71, 1.76), in the direct-effect model, the residual effect of age is below 1 after controlling for length and \PT\ (0.90). This means that at the same length and the same \PT\ condition, an older child can receive a lower \Ind\ score. This aligns with the scoring design of Independence, where the upper levels from 5 to 10 impose no minimum length requirement and instead assess discourse functions such as managing the flow of conversation or shifting topics. Thus, it is not how long the child speaks but how the child responds to the prompt and which discourse move the child selects that determines the score. The fact that a negative residual age slope remains even after holding length and \PT\ fixed shows that differences in strategy and context sensitivity exist among children even at the same length.


\subsection{Age Prediction}
\label{sec:agepred}
\begin{table*}[ht]
\centering
\small
\setlength{\tabcolsep}{4pt}
\begin{adjustbox}{max width=\textwidth}
\begin{tabular}{@{}ll ccc ccc ccc ccc ccc ccc@{}}
\toprule
 &  & \multicolumn{6}{c}{\textbf{Mixtral-8x7B-Inst.}} & \multicolumn{6}{c}{\textbf{Gemini 2.5 Pro}} \\
\cmidrule(lr){3-8}\cmidrule(lr){9-14}
\textbf{Type}& \textbf{Feature set} & \multicolumn{2}{c}{\textbf{MAE}} & \multicolumn{2}{c}{\textbf{MSE}} & \multicolumn{2}{c}{$\mathbf{R^2}$} 
& \multicolumn{2}{c}{\textbf{MAE}} & \multicolumn{2}{c}{\textbf{MSE}} & \multicolumn{2}{c}{$\mathbf{R^2}$} \\
\midrule
LLM & Child utterances & \multicolumn{2}{c}{1.30} & \multicolumn{2}{c}{2.75} & \multicolumn{2}{c}{0.51} & \multicolumn{2}{c}{1.03} & \multicolumn{2}{c}{2.25} & \multicolumn{2}{c}{0.51} \\
\midrule
& & \multicolumn{3}{c}{\textbf{Linear Regression}} & \multicolumn{3}{c}{
\textbf{Gradient Boosting}} & \multicolumn{3}{c}{\textbf{Linear Regression}} & \multicolumn{3}{c}{\textbf{Gradient Boosting}} \\
\cmidrule(lr){3-5}\cmidrule(lr){6-8}\cmidrule(lr){9-11}\cmidrule(lr){12-14}
& & \textbf{MAE} & \textbf{MSE} & $\mathbf{R^2}$ & \textbf{MAE} & \textbf{MSE} & $\mathbf{R^2}$ & \textbf{MAE} & \textbf{MSE} & $\mathbf{R^2}$ & \textbf{MAE} & \textbf{MSE} & $\mathbf{R^2}$ \\
\midrule
\multirow{8}{*}{Baselines} & MLU & 1.53 & 3.84 & 0.17 & 1.52 & 4.25 & 0.07 & 1.42 & 3.47 & 0.23 & 1.55 & 4.10 & 0.09 \\
& MLU + \PT & 1.51 & 3.78 & 0.18 & 1.50 & 4.15 & 0.09 & 1.40 & 3.41 & 0.25 & 1.50 & 3.91 & 0.14 \\
& vocd\text{-}D & 1.76 & 4.58 & 0.01 & 1.81 & 4.87 & -0.06 & 1.75 & 4.55 & 0.00 & 1.78 & 4.66 & -0.03 \\
& vocd\text{-}D + \PT & 1.72 & 4.47 & 0.03 & 1.76 & 4.63 & 0.00 & 1.71 & 4.41 & 0.03 & 1.70 & 4.35 & 0.04 \\
& FKGL & 1.76 & 4.55 & 0.02 & 1.73 & 4.56 & 0.01 & 1.72 & 4.40 & 0.03 & 1.67 & 4.27 & 0.06 \\
& FKGL + \PT & 1.73 & 4.46 & 0.03 & 1.71 & 4.47 & 0.03 & 1.69 & 4.31 & 0.05 & 1.63 & 4.13 & 0.09 \\
& GFI & 1.66 & 4.22 & 0.09 & 1.52 & 3.80 & 0.18 & 1.57 & 3.92 & 0.14 & 1.43 & 3.43 & 0.25 \\
& GFI + \PT & 1.64 & 4.17 & 0.10 & 1.52 & 3.75 & 0.19 & 1.56 & 3.87 & 0.15 & 1.40 & 3.31 & 0.27 \\
\midrule
\multirow{6}{*}{Ours} & \Exp\ + \Ind & 1.20 & 2.75 & 0.40 & 1.23 & 2.97 & 0.36 & 1.14 & 2.41 & 0.47 & 1.05 & 2.09 & 0.54 \\
& \Exp\ + \Ind\ + \PT & \textbf{1.19}$^{\ast}$ & \textbf{2.71}$^{\ast}$ & \textbf{0.41}$^{\ast}$ & \textbf{1.22}$^{\ast}$ & \textbf{2.89}$^{\ast}$ & \textbf{0.37}$^{\ast}$ & \textbf{1.13}$^{\ast}$ & \textbf{2.38}$^{\ast}$ & \textbf{0.48}$^{\ast}$ & \textbf{1.04}$^{\ast}$ & \textbf{2.06}$^{\ast}$ & \textbf{0.55}$^{\ast}$ \\
& \Exp & 1.20 & 2.75 & 0.40 & 1.23 & 2.97 & 0.36 & 1.22 & 2.76 & 0.39 & 1.14 & 2.47 & 0.46 \\
& \Exp\ + \PT & 1.20 & 2.73 & 0.41 & 1.23 & 2.97 & 0.36 & 1.21 & 2.71 & 0.40 & 1.12 & 2.42 & 0.47 \\
& \Ind & 1.34 & 3.17 & 0.31 & 1.35 & 3.38 & 0.26 & 1.24 & 2.76 & 0.39 & 1.33 & 3.03 & 0.33 \\
& \Ind\ + \PT & 1.33 & 3.13 & 0.32 & 1.34 & 3.33 & 0.27 & 1.23 & 2.72 & 0.40 & 1.33 & 3.03 & 0.33 \\
\bottomrule
\end{tabular}
\end{adjustbox}
\caption{Age prediction with child-grouped 5-fold CV. ($^{\ast}$) indicates the best performance. 95\% confidence intervals (CIs) are reported in Appendix Tables~\ref{apptab:cis_mixtral} and~\ref{apptab:cis_gemini}.}
\label{tab:agepred}
\end{table*}

This section validates whether the developmental pathway decomposition confirmed in Section~\ref{sec:dev-validity} generalizes to age prediction, not merely providing explanatory power. We assess whether our metrics provide additional predictive power beyond baseline metrics (MLU, vocd-D, FKGL, GFI), demonstrating that the developmental signal carries predictive validity.

\paragraph{Setup} 
All experiments use only child utterances as input, with the target variable being the child's age in decimal years (e.g., 4 years 7 months = 4.58 years). We employed GroupKFold (k=5) to ensure all utterances from the same child remain in the same fold, preventing data leakage. We report mean absolute error, mean squared error, coefficient of determination $R^2$, and 95\% confidence intervals. FKGL, GFI, and vocd-D, which require sufficient text length for stable estimation, were computed by concatenating all utterances for each child. We evaluate both with and without \PT\ features. For our metrics, which are ordinal and distributional by design, we summarize E/I per child using Low/Medium/High ratios (1-3/4-7/8-10), and also compute \PT-wise macro means and standard deviations to reflect context sensitivity (details in Appendix~\ref{app:agepred}). We also include an LLM baseline using open-weight model \verb|Mixtral-8x7B-Instruct-v0.1| (chosen for its efficient MoE architecture and 32K context) and closed-weight model \verb|Gemini 2.5 Pro| that directly predict age from child utterances only. Additionally, \verb|Mixtral| labels PT, Expansion, and Independence features for linear/non-linear models.


\paragraph{Results} 
Table~\ref{tab:agepred} shows that E/I features outperform all baselines under both Linear Regression (LR) and Gradient Boosting (GB). Under LR, the best baseline is MLU with \PT\ (1.40), whereas Ours (\textit{E} + \textit{I} + \PT) attains a 19.3\% MAE reduction. Under GB, the reduction is even larger at 25.7\% compared to the best baseline GFI with \PT. 
Although Section~\ref{sec:dev-validity} shows that vocd-D increases with age ($\beta \approx$16-17), the actual age ranges overlap widely at the same vocd-D value, yielding low $R^2$ and the largest MAE in prediction. This arises because vocd-D exhibits $\beta_{b} < 0$ (Table ~\ref{tab:trad-vs-ours}), reflecting its sensitivity to habitual verbosity at the between-child level.
We also observe sizeable gains in association with age, as $R^2$ improves by +0.23 in Linear (0.25→0.48) and +0.28 in GB (0.27→0.55) over the strongest baselines.
While adding \PT\ improves all feature sets, the gain is comparatively smaller for our \Exp/\Ind\ features than for traditional baselines, suggesting that our summary statistics already capture \PT-specific effects. 
Analyzing individual contributions of \Exp\ and \Ind, \Exp-only showed stronger predictive power than \Ind-only.
This aligns with the findings in Section~\ref{sec:dev-validity} that Expansion is connected to age through the length-mediated pathway. However, combining both metrics yielded the best performance, confirming they provide complementary developmental signals that generalize beyond explanation to prediction.

When we evaluate simple age prediction with LLMs using only children's utterances, \verb|Mixtral| and \verb|Gemini| attain MAEs of approximately 1.3 and 1.03, respectively, predicting age in decimal years (conversion details in Appendix~\ref{app:dataset}). Predicting age down to the month level is a challenging task requiring fine-grained discrimination. Given these low errors, we cannot rule out contamination from potential pretraining exposure to the CHILDES dataset (we present supporting checks in Appendix~\ref{app:llmconta}). Therefore, because the low errors on CHILDES may reflect prior exposure during pretraining, we cannot ascertain the models' true ability to predict age, and it remains uncertain whether they would generalize beyond CHILDES. In contrast, our approach leverages general developmental concepts rather than memorizing dataset-specific patterns, and thus should be more robust.



\begin{table*}[ht]
\centering
\small
\setlength{\tabcolsep}{7pt}
\begin{tabular}{l c c c c c}
\toprule
\textbf{Baseline (LMM)} & \textbf{Model} & \textbf{Effect Type}
& \textbf{Causal}
& \textbf{Contrast}
& \textbf{Initiative} \\
\midrule
\multirow{2}{*}{MLU}
& S1 & \multirow{2}{*}{$\beta$} 
& $\approx 0$ ($P>.10$) 
& $\approx 0$ ($P>.10$) 
& $\approx 0$ ($P>.10$) \\
& S2 & 
& \textbf{3.03} ($P<.001$) 
& \textbf{3.42} ($P<.001$) 
& \textbf{1.98} ($P<.001$) \\
\midrule
\multicolumn{6}{l}{\textbf{Our metrics (CLMM)}}\\
\midrule
\multirow{2}{*}{Expansion (\Exp)}
& S1 & \multirow{2}{*}{OR} 
& \textbf{12.10} ($P<.001$) 
& \textbf{6.73} ($P<.001$) 
& \textbf{3.44} ($P<.001$) \\
& S2 & 
& \textbf{4.49} ($P<.001$) 
& \textbf{2.36} ($P<.001$) 
& \textbf{1.86} ($P<.001$) \\
\addlinespace[2pt]
\multirow{2}{*}{Independence (\Ind)}
& S1 & \multirow{2}{*}{OR} 
& \textbf{2.54} ($P<.001$) 
& \textbf{1.77} ($P<.001$) 
& \textbf{3.85} ($P<.001$) \\
& S2 & 
& \textbf{2.11} ($P<.001$) 
& \textbf{1.84} ($P<.001$) 
& \textbf{4.00} ($P<.001$) \\
\bottomrule
\end{tabular}
\caption{Comparison of baselines and our metrics across discourse marker types. All analyses use \PT\ and \Exp/\Ind\ scores annotated by Gemini 2.5 Pro. Effects that are significant are bolded ($P<.001$); non-significant effects are shown as $P>.10$.}
\label{tab:metrics-vs-ours-semantic}
\end{table*}

\subsection{Semantic Sensitivity}
\label{sec:sem-sensitivity}
Section~\ref{sec:dev-validity} decomposed age effects and showed that Expansion is length-mediated while Independence reflects contextual dependence. This section examines whether our metrics are semantically sensitive when length is not a factor. Specifically, we ask the following question: 
Within utterances matched on length and \PT, do Expansion and Independence vary with discourse structure (e.g., causality, contrast, initiative) in ways that a length baseline such as MLU does not? By isolating semantic/discourse markers under fixed length and \PT, we test our metrics' ability to capture meaning-level differences beyond length, complementing the developmental validity in Section~\ref{sec:dev-validity}. 

\paragraph{Setup} 
In Equation~\ref{eq:s1} (S1), we hold length and \PT\ fixed to isolate and test the pure semantic effect of the presence of discourse markers $M_k$ on the scores. This design allows us to attribute any remaining differences purely to the presence or absence of markers (see Appendix Table~\ref{apptab:markers-to-rubric}). 
We discretize utterance length (in words) into three bands, \(\textit{LenBin} \in \{1\text{-}3,\,4\text{-}6,\,7\text{-}10\}\). 
In Equation~\ref{eq:s2} (S2), we deliberately relax the cell-wise constraint to test robustness under mixed lengths. Equations ~\ref{eq:s1},~\ref{eq:s2} define the shared linear predictor $\eta$.  \footref{fn:lmm}


\begin{equation}
\label{eq:s1}
Y 
= \beta_0 + \sum_k \beta_{M_k} M_k
+ u_{\text{child}}
+ u_{\text{\PT:\textit{LenBin}}} + \varepsilon
\end{equation}

\begin{equation}
\label{eq:s2}
\begin{aligned}
Y = \beta_0 + & \sum_k \beta_{M_k} M_k
+ \beta_w L^{(w)}_z + \beta_b L^{(b)}_z \\&
+ \boldsymbol{\beta}_{\!PT}^{\top}PT
+ u_{\text{child}} + \varepsilon
\end{aligned}
\end{equation}

If odds ratios that are large in S1 remain clearly $>1$ and significant in S2, the signal is not an artifact of cell conditioning but genuinely survives under realistic length variation. Some attenuation is expected because length heterogeneity is reintroduced.
We exclude vocd-D, FKGL, and GFI, as these measures assume concatenation or sufficient text for stable estimation. 

\paragraph{Analysis} 
For \Exp\ in Table~\ref{tab:metrics-vs-ours-semantic}, causal markers yield an OR of approximately 12.10, meaning that within the same length–\textit{PT} cell, the presence of causal connectives increases the odds of moving to a higher category by roughly twelvefold. 
In S2, this attenuates to 4.49, but the direction and significance remain, indicating that causal linking boosts perceived \Exp\ even after reintroducing length heterogeneity. For \Ind, causal markers yield ORs of 2.54 in S1 and 2.11 in S2, both clearly $>1$, suggesting that causality reflects more than verbosity. It aligns with the child's initiative in advancing discourse. 
Similarly, contrast markers raise both \Exp\ and \Ind\ at equal length in S1 (\(\Exp \approx 6.7\times\), \(\Ind \approx 1.8\times\)), and their effects remain substantial in S2.
Initiative markers are positive for \Exp\ as well, but their impact is strongest on \Ind\ (\(\approx 4.0\times\)) and remains robust in S2, consistent with the idea that proposals or tag questions strongly signal discourse advancement. Turning to MLU, in S1, it does not respond to marker presence at all, as expected when length is fixed within cells. In S2, positive coefficients emerge, but this reflects a side effect of length increase rather than sensitivity to meaning, as utterances with such markers tend to be longer when length is not constrained. In other words, MLU responds to utterance length but not to semantic content or discourse strategy, functioning as a length-dependent proxy rather than a measure of meaning.

\subsection{Human Evaluation}
\label{app:humeval}
We evaluate how well the proposed Expansion and Independence scores align with human judgments.
From the adult--child dialogues, we draw a stratified sample of 300 turns and obtain annotations from three independent expert raters using the same rubric as the LLM.
To prevent anchoring bias, raters did not have access to the models' predictions.
We report human--human inter-annotator agreement (Table~\ref{tab:human-human-iaa}) to assess the reliability of the rubric using Krippendorff's $\alpha$. Krippendorff's $\alpha$ values (Table~\ref{tab:human-human-iaa}) suggest substantial agreement for Expansion and Independence, and moderate agreement for PT.
For model--human alignment, we do not treat machine labels as interchangeable with human annotations. Instead, we evaluate the model’s ability to reproduce human ratings using an error-sensitive metric appropriate for ordinal targets (average MAE to each rater), following recommendations for ordinal evaluation \citep{chen-etal-2021-error} and prior work cautioning against agreement measures that assume interchangeable annotators for machine-generated labels \citep{elangovan2025correlationimpacthumanuncertainty}. This yields MAE = 0.95/1.12 (Expansion/Independence) for Gemini 2.5 Pro and 1.16/1.45 for Mixtral-8x7B-Instruct.


\begin{table}[ht]
\centering
\begin{adjustbox}{max width=\columnwidth}
\begin{tabular}{lccc}
\toprule
\textbf{Metric} 
& \textbf{Krippendorff's $\boldsymbol\alpha$} 
& \textbf{Fleiss' $\boldsymbol\kappa$} 
& \textbf{Avg Cohen's $\boldsymbol\kappa$} \\
\midrule
PT            & 0.602 {\color{gray}[0.556, 0.647]} & 0.601 {\color{gray}[0.555, 0.646]} & 0.601 {\color{gray}[0.556, 0.647]} \\
Expansion     & 0.858 {\color{gray}[0.827, 0.880]} & -- & -- \\
Independence  & 0.759 {\color{gray}[0.701, 0.809]} & -- & -- \\
\bottomrule
\end{tabular}
\end{adjustbox}
\caption{Human--human inter-annotator agreement among three expert raters. We report Krippendorff's $\alpha$ for ordinal tasks (Expansion/Independence) and $\alpha/\kappa$ for nominal PT. Brackets indicate 95\% confidence intervals.}
\label{tab:human-human-iaa}
\end{table}

\section{Broader Impacts}
\paragraph{Pedagogical Evaluation and Tutoring Optimization}
Our metrics provide pedagogically meaningful signals for evaluating and optimizing educational dialogue systems. Unlike standard overlap-based metrics such as BLEU and ROUGE, which encourage spoon-feeding behaviors, our framework evaluates interactional outcomes by measuring whether a system elicits more Expanded and Independent responses from learners. In education, the goal is not merely to provide correct answers, but to ask questions that encourage learners to think. For example, scaffolding questions that prompt reasoning are often pedagogically preferable to direct answers, yet are penalized by standard metrics for low lexical overlap with reference responses. In contrast, our approach focuses on interactional outcomes rather than surface form. As shown in Table~\ref{tab:exp-indep-summary}, the proposed metrics capture systematic differences between questioning strategies (e.g., open vs.\ closed prompts). For instance, a tutoring agent could be rewarded for using Elaboration-type prompts that elicit richer child contributions (median Expansion = 4), rather than relying on constrained formats such as Yes/No or Display questions (median Expansion = 2, Independence IQR = 0), which limit opportunities for autonomous child talk.
Moreover, they can serve as dense, turn-level reward signals for reinforcement learning, enabling tutoring agents to prioritize scaffolding behaviors such as guided questioning rather than overly verbose instruction.

\paragraph{Auditing Synthetic Dialogues and AI Safety}
With rising interest in simulating children \citep{Milika2024LargeLM}, recent work has also focused on constructing and analyzing synthetic child–adult dialogues \citep{feng2024childdirectedspeecheffectivetraining}. 
However, existing studies primarily focus on task correctness and do not verify whether generated conversations reflect developmentally natural discourse. This raises a critical question of who evaluates whether synthetic data captures authentic child discourse characteristics. We argue that our metrics provide a promising approach for this essential quality control and auditing role. By verifying whether synthetic dialogues adhere to age-appropriate Expansion and Independence distributions, researchers can filter unrealistic mimicry and ensure the quality of training data.

Beyond data auditing, our results also have implications for AI safety and age assurance. AI providers (e.g., OpenAI's age prediction) actively seek robust age assurance methods to enforce safety policies. We demonstrate a strong correlation between discourse-level metrics and age (Section~\ref{sec:agepred}), suggesting that these interpretable linguistic features could complement keyword-based filters by enabling platforms to distinguish child users based on language maturity rather than surface lexical cues.

\section{Conclusion}
We introduced two context-aware metrics, \textbf{Expansion} and \textbf{Independence}, that assess how a child's utterance organizes meaning and advances discourse, not just how long it is. Conditioning on the \textbf{Previous Adult Utterance Type}, \Exp/\Ind\ show clear distributions, remain developmentally valid, capture semantic/discourse structure, and improve age prediction over baselines. Our metrics refocus evaluation on context-sensitive discourse and enable multimodal and cross-linguistic extensions.


\section*{Limitations}
We analyze orthographic transcripts only, so prosody and acoustic cues are not captured and are left for future work.
Our results come from North American English corpora and may not generalize to other languages. For the LLM baselines when predicting age, the models may have encountered highly similar conversations during pretraining.

\section*{Ethical Considerations}
We use publicly available, de-identified child–adult transcripts (CHILDES/TalkBank) under their terms of use and make no attempt to re-identify individuals. 
Age-related analyses are used to validate the developmental sensitivity of our metrics, not to infer or verify the identity of individual users.
The metrics are not intended for diagnostic, clinical, or high-stakes decisions about children. LLM-as-a-judge scoring may encode societal or dataset biases. To address this, we release prompts/code, report rubric details, and encourage independent replication and auditing. We avoid releasing any sensitive raw text beyond what licenses permit, and we caution against using these metrics to gate access or impose labels on children in real-world systems.

\section*{Acknowledgments}
This work was partially supported by a Battelle Engineering, Technology and Human Affairs (BETHA) Endowment grant and by a CCBS Summer Graduate Research Award from the Center for Cognitive and Brain Sciences. We also gratefully acknowledge the computational resources provided by the Ohio Supercomputer Center (OSC).


\bibliography{custom}
\appendix

\section{Previous Adult Utterance Type Specification and Related Research}
\label{app:pt}

\paragraph{Supporting Literature}
We define Previous Adult Utterance Type (\PT) as a taxonomy of adult utterances,
covering both \emph{question} and \emph{non-question} categories based on the adult's 
communicative intent (e.g., eliciting information vs.\ providing feedback).
Our taxonomy draws on constructs from prior work in
parent--child interaction and dialogue-act research, where adult
question types and scaffolding strategies have been shown to influence children's
responses and language development. Table~\ref{apptab:pt-mapping}
lists our categories with representative citations.

\begin{table*}[t]
\centering
\small
\setlength{\tabcolsep}{7pt}
\begin{tabular}{l l}
\toprule
\textbf{\PT} & \textbf{Representative citations} \\
\midrule
Display Question            & \citet{Display} \\
Yes/No Question             & \citet{stolcke-etal-2000-dialogue, Display} \\
Choice Question             & \citet{museum, callanan2020exploration} \\
Referential Question        & \citet{wh, Display} \\
Clarification Request       & \citet{Clark02072020} \\
Elaboration Prompt          & \citet{fivush2006elaborating, callanan2020exploration} \\
Confirmation Check          & \citet{Clark02072020, leech2013father} \\
\midrule
Topic Introduction          & \citet{callanan2020exploration, Rowe2019AnalyzingIQ} \\
Imperative/Request          & \citet{stolcke-etal-2000-dialogue} \\
Feedback/Evaluation         & \citet{Rowe2012ALI, sp} \\
Self-Talk/Thinking Aloud    & \citet{stolcke-etal-2000-dialogue} \\
\bottomrule
\end{tabular}
\caption{Previous Adult Utterance Type (\PT) taxonomy and representative prior work.}
\label{apptab:pt-mapping}
\end{table*}

\paragraph{Prompt}
To support reproducibility, we provide the full system prompt used for LLM-based classification of adult turns (Table~\ref{apptab:classification_prompt}).
The prompt assigns a single, mutually exclusive label to each adult utterance based on its primary communicative intent. It determines whether an utterance is a question from content and structure rather than punctuation, and applies question and non-question taxonomies grounded in prior work on parent--child interaction and conversational scaffolding.
Ambiguous or unintelligible turns receive a special label to prevent spurious classifications.

\section{Expansion and Independence Specification and Related Research}
\label{app:EI}

\paragraph{Supporting Literature}
We define Independence (\Ind) and Expansion (\Exp) as complementary dimensions of child language, grounded in prior developmental research. 
\Ind~measures conversational agency from imitation to pragmatic leadership, while \Exp~tracks utterance elaboration from single words to complex narratives.
Tables~\ref{apptab:expansion-mapping} and~\ref{apptab:independence-mapping} list the foundational studies underlying each score level.

\paragraph{Prompt}
We provide the full system prompt used for LLM-based scoring of child utterances on Independence and Expansion (Table~\ref{apptab:scoring_prompt}). 
The prompt first filters out hesitation-only responses, then scores content-based utterances.

\section{Dataset Preprocessing}
\label{app:dataset}
We use 20 corpora from CHILDES: \textit{Bliss, Bohannon, Braunwald, Brown, Champaign, Clark, Demetras1, Demetras2, EllisWeismer, Gleason, HSLLD, Kuczaj, McCune, Nelson, NewmanRatner, OCSC, Post, Sachs, Suppes, Warren}. 
For each corpus, we extract child and adult utterances from the transcripts, align them by speaker turns, and pass them to the analysis pipeline.

\paragraph{Transcript cleaning and tokenization}
We apply three preprocessing steps. 
First, we retain standard CHILDES markup while removing symbols irrelevant to analysis, including tokens consisting only of \texttt{xxx} (marking unintelligible speech).
Second, for underscore-compounded words (\texttt{under\_score\_word}), we replace underscores with spaces to enable space-based tokenization (e.g., \texttt{under\_score\_word} $\rightarrow$ \texttt{under score word}).

\paragraph{Session and utterance filtering}
We remove segments that provide weak predictive signal or may introduce structural bias:
\begin{itemize}
  \item \textbf{Memorized or recited material}: songs, chants, verse recitation, alphabet or number reading
  \item \textbf{Read-aloud or elicited repetition}: shared book reading, read-aloud tasks, explicit repetition prompts
  \item \textbf{Word-list production}: picture naming, word lists, nonword or syllable repetition
  \item \textbf{Fixed scripts}: play-acting driven by scripted dialogue
\end{itemize}
We also exclude empty lines, non-utterance lines, and lines containing only noise markers such as \texttt{xxx}. 
Utterances consisting solely of hesitation or filler sounds (\textit{um, uh, hm}) are treated as \textit{hesitation-only}, assigned Expansion=0, Independence=0, and then excluded from analysis.

\paragraph{Age conversion}
Age notations in transcripts (\texttt{P\#Y\#M}, \texttt{6;9.12}, and related formats) are converted to decimal years. For example, \texttt{P6Y07M} $\rightarrow$ 6.58 years and \texttt{4;7.0} $\rightarrow$ 4.58 years.

\section{Discourse Markers}
\label{app:Discourse}

We use a set of surface-level discourse markers to categorize pragmatic and structural cues in adult utterances.
These markers guide how child responses are scored for Expansion and Independence by capturing cues such as causal reasoning, contrastive organization, and conversational initiative.
Table~\ref{apptab:markers-to-rubric} lists the marker categories and corresponding lexical patterns.


\section{Age Prediction}
\label{app:agepred}

\paragraph{LLM Age Prediction Prompt}
The following system prompt (Table~\ref{apptab:llm-age-prompt}) was used for the large language model evaluation described in Section~\ref{sec:agepred}. 
The model inferred a child's age in years and months from a transcript containing only the child's utterances.

\begin{table*}[ht]
\centering
\small
\setlength{\tabcolsep}{7pt}
\begin{tabular}{c l}
\toprule
\textbf{Expansion score} & \textbf{Representative citations} \\
\midrule
1  & \citet{BLOOM1976521} \\
2  & \citet{Braine1963_Pivot, Radford1990_Telegraphic} \\
3  & \citet{Radford1990_Telegraphic, BLOOM1976521} \\
4  & \citet{clauses} \\
5  & \citet{clauses, Rowe2019AnalyzingIQ} \\
6  & \citet{Kallay, clauses} \\
7  & \citet{EVERSVERMEUL20111645, Maternal_Elaborative_Reminiscence_Mediation} \\
8  & \citet{McCabe1996RelatingEI, fivush2006elaborating, Narrative} \\
9  & \citet{clauses, EVERSVERMEUL20111645} \\
10 & \citet{McCabe1996RelatingEI, fivush2006elaborating, clauses} \\
\bottomrule
\end{tabular}
\caption{Mapping of \Exp~scores (1--10) to representative prior work (no MLU citations).}
\label{apptab:expansion-mapping}
\end{table*}

\begin{table*}[ht]
\centering
\small
\setlength{\tabcolsep}{7pt}
\begin{tabular}{c l}
\toprule
\textbf{Independence score} & \textbf{Representative citations} \\
\midrule
1  & \citet{BLOOM1976521} \\
2  & \citet{BLOOM1976521} \\
3  & \citet{Clark02072020, YesNo_Action_Tendencies} \\
4  & \citet{Display, SALOMO_LIEVEN_TOMASELLO_2013, YesNo_Action_Tendencies} \\
5  & \citet{Clark02072020, leech2013father} \\
6  & \citet{Chouinard2007ChildrensQA, wh} \\
7  & \citet{fivush2006elaborating, Rowe2019AnalyzingIQ, Maternal_Elaborative_Reminiscence_Mediation} \\
8  & \citet{McCabe1996RelatingEI, fivush2006elaborating} \\
9  & \citet{Inde, Clark02072020, VANDAM2022101338} \\
10 & \citet{MASEK2021100961, Inde} \\
\bottomrule
\end{tabular}
\caption{Mapping of \Ind~scores (1--10) to representative prior work (no MLU citations).}
\label{apptab:independence-mapping}
\end{table*}
\paragraph{Training Details}
All models were trained using 5-fold GroupKFold cross-validation to prevent child-level data leakage. Hyperparameters were selected based on validation MAE. 
For each fold, we computed mean absolute error (MAE), mean squared error (MSE), 
and coefficient of determination ($R^2$), reporting mean values and 95\% confidence intervals.
For the non-linear model, we used scikit-learn's \texttt{GradientBoostingRegressor} with 500 estimators, a learning rate of 0.05, maximum depth of 3, and a fixed random state.
\paragraph{95\% CIs}
Tables~\ref{apptab:cis_mixtral} and~\ref{apptab:cis_gemini} show 95\% confidence intervals (CIs).

\begin{table*}[ht]
\centering
\small
\setlength{\tabcolsep}{6pt}
\begin{adjustbox}{max width=\textwidth}
\begin{tabular}{@{}ll ccc ccc@{}}
\toprule
\textbf{Type} & \textbf{Feature set} 
& \multicolumn{3}{c}{\textbf{Linear Regression}} 
& \multicolumn{3}{c}{\textbf{Gradient Boosting}} \\
\cmidrule(lr){3-5}\cmidrule(lr){6-8}
& & \textbf{MAE} & \textbf{MSE} & $\mathbf{R^2}$ 
& \textbf{MAE} & \textbf{MSE} & $\mathbf{R^2}$ \\
\midrule

\multirow{8}{*}{Baselines}
& MLU 
& 1.53 {\color{blue}[1.39–1.68]} & 3.84 & 0.17 {\color{blue}[0.12–0.21]}
& 1.52 {\color{blue}[1.34–1.70]} & 4.25 & 0.07 {\color{blue}[-0.28–0.38]} \\
& MLU + \PT 
& 1.51 {\color{blue}[1.36–1.66]} & 3.78 & 0.18 {\color{blue}[0.14–0.22]}
& 1.50 {\color{blue}[1.32–1.69]} & 4.15 & 0.09 {\color{blue}[-0.25–0.38]} \\
& vocd\text{-}D 
& 1.76 {\color{blue}[1.65–1.87]} & 4.58 & 0.01 {\color{blue}[0.00–0.02]}
& 1.81 {\color{blue}[1.73–1.90]} & 4.87 & -0.06 {\color{blue}[-0.12–0.00]} \\
& vocd\text{-}D + \PT 
& 1.72 {\color{blue}[1.61–1.84]} & 4.47 & 0.03 {\color{blue}[0.02–0.04]}
& 1.76 {\color{blue}[1.66–1.86]} & 4.63 & 0.00 {\color{blue}[-0.05–0.03]} \\
& FKGL 
& 1.76 {\color{blue}[1.61–1.84]} & 4.55 & 0.02 {\color{blue}[0.02–0.04]}
& 1.73 {\color{blue}[1.57–1.87]} & 4.56 & 0.01 {\color{blue}[-0.03–0.07]} \\
& FKGL + \PT 
& 1.73 {\color{blue}[1.61–1.84]} & 4.46 & 0.03 {\color{blue}[0.02–0.04]}
& 1.71 {\color{blue}[1.57–1.85]} & 4.47 & 0.03 {\color{blue}[-0.03–0.07]} \\
& GFI 
& 1.66 {\color{blue}[1.54–1.78]} & 4.22 & 0.09 {\color{blue}[0.07–0.11]}
& 1.52 {\color{blue}[1.39–1.64]} & 3.80 & 0.18 {\color{blue}[0.15–0.22]} \\
& GFI + \PT 
& 1.64 {\color{blue}[1.52–1.76]} & 4.17 & 0.10 {\color{blue}[0.08–0.12]}
& 1.52 {\color{blue}[1.39–1.64]} & 3.75 & 0.19 {\color{blue}[0.15–0.22]} \\
\midrule

\multirow{6}{*}{Ours}
& \Exp\ + \Ind 
& 1.20 {\color{blue}[1.11–1.27]} & 2.75 & 0.40 {\color{blue}[0.36–0.46]}
& 1.23 {\color{blue}[1.12–1.31]} & 2.97 & 0.36 {\color{blue}[0.29–0.45]} \\
& \Exp\ + \Ind\ + \PT 
& \textbf{1.19}$^{\ast}$ {\color{blue}[1.12–1.28]} & \textbf{2.71}$^{\ast}$ & \textbf{0.41}$^{\ast}$ {\color{blue}[0.35–0.46]}
& \textbf{1.22}$^{\ast}$ {\color{blue}[1.12–1.31]} & \textbf{2.89}$^{\ast}$ & \textbf{0.37}$^{\ast}$ {\color{blue}[0.29–0.45]} \\
& \Exp 
& 1.20 {\color{blue}[1.12–1.28]} & 2.75 & 0.40 {\color{blue}[0.35–0.46]}
& 1.23 {\color{blue}[1.09–1.38]} & 2.97 & 0.36 {\color{blue}[0.29–0.42]} \\
& \Exp\ + \PT 
& 1.20 {\color{blue}[1.11–1.28]} & 2.73 & 0.41 {\color{blue}[0.35–0.46]}
& 1.23 {\color{blue}[1.09–1.37]} & 2.97 & 0.36 {\color{blue}[0.29–0.41]} \\
& \Ind 
& 1.34 {\color{blue}[1.25–1.43]} & 3.17 & 0.31 {\color{blue}[0.27–0.36]}
& 1.35 {\color{blue}[1.23–1.48]} & 3.38 & 0.26 {\color{blue}[0.15–0.36]} \\
& \Ind\ + \PT 
& 1.33 {\color{blue}[1.24–1.42]} & 3.13 & 0.32 {\color{blue}[0.28–0.36]}
& 1.34 {\color{blue}[1.23–1.47]} & 3.33 & 0.27 {\color{blue}[0.16–0.37]} \\
\bottomrule
\end{tabular}
\end{adjustbox}
\caption{Mixtral-8x7B-Instruct results. 95\% confidence intervals (CIs, shown in blue) are reported.}
\label{apptab:cis_mixtral}
\end{table*}

\begin{table*}[ht]
\centering
\small
\setlength{\tabcolsep}{6pt}
\begin{adjustbox}{max width=\textwidth}
\begin{tabular}{@{}ll ccc ccc@{}}
\toprule
\textbf{Type} & \textbf{Feature set} 
& \multicolumn{3}{c}{\textbf{Linear Regression}} 
& \multicolumn{3}{c}{\textbf{Gradient Boosting}} \\
\cmidrule(lr){3-5}\cmidrule(lr){6-8}
& & \textbf{MAE} & \textbf{MSE} & $\mathbf{R^2}$ 
& \textbf{MAE} & \textbf{MSE} & $\mathbf{R^2}$ \\
\midrule

\multirow{8}{*}{Baselines}
& MLU 
& 1.42 {\color{blue}[1.36–1.48]} & 3.47 & 0.23 {\color{blue}[0.15–0.31]}
& 1.55 {\color{blue}[1.44–1.66]} & 4.10 & 0.09 {\color{blue}[0.08–0.22]} \\
& MLU + \PT 
& 1.40 {\color{blue}[1.35–1.46]} & 3.41 & 0.25 {\color{blue}[0.17–0.32]}
& 1.50 {\color{blue}[1.41–1.55]} & 3.91 & 0.14 {\color{blue}[0.11–0.25]} \\
& vocd\text{-}D 
& 1.75 {\color{blue}[1.67–1.83]} & 4.55 & 0.00 {\color{blue}[-0.01–0.00]}
& 1.78 {\color{blue}[1.73–1.82]} & 4.66 & -0.03 {\color{blue}[-0.06–0.02]} \\
& vocd\text{-}D + \PT 
& 1.71 {\color{blue}[1.63–1.79]} & 4.41 & 0.03 {\color{blue}[0.02–0.03]}
& 1.70 {\color{blue}[1.64–1.74]} & 4.35 & 0.04 {\color{blue}[0.00–0.09]} \\
& FKGL 
& 1.72 {\color{blue}[1.64–1.79]} & 4.40 & 0.03 {\color{blue}[0.02–0.04]}
& 1.67 {\color{blue}[1.59–1.75]} & 4.27 & 0.06 {\color{blue}[0.01–0.10]} \\
& FKGL + \PT 
& 1.69 {\color{blue}[1.61–1.76]} & 4.31 & 0.05 {\color{blue}[0.04–0.06]}
& 1.63 {\color{blue}[1.56–1.72]} & 4.13 & 0.09 {\color{blue}[0.03–0.12]} \\
& GFI 
& 1.57 {\color{blue}[1.49–1.65]} & 3.92 & 0.14 {\color{blue}[0.10–0.17]}
& 1.43 {\color{blue}[1.38–1.49]} & 3.43 & 0.25 {\color{blue}[0.19–0.29]} \\
& GFI + \PT 
& 1.56 {\color{blue}[1.48–1.64]} & 3.87 & 0.15 {\color{blue}[0.11–0.18]}
& 1.40 {\color{blue}[1.34–1.46]} & 3.31 & 0.27 {\color{blue}[0.22–0.31]} \\
\midrule

\multirow{6}{*}{Ours}
& \Exp\ + \Ind 
& 1.14 {\color{blue}[1.11–1.17]} & 2.41 & 0.47 {\color{blue}[0.43–0.51]}
& 1.05 {\color{blue}[0.99–1.06]} & 2.09 & 0.54 {\color{blue}[0.49–0.61]} \\
& \Exp\ + \Ind\ + \PT 
& \textbf{1.13}$^{\ast}$ {\color{blue}[1.10–1.16]} & \textbf{2.38}$^{\ast}$ & \textbf{0.48}$^{\ast}$ {\color{blue}[0.44–0.51]}
& \textbf{1.04}$^{\ast}$ {\color{blue}[0.98–1.05]} & \textbf{2.06}$^{\ast}$ & \textbf{0.55}$^{\ast}$ {\color{blue}[0.50–0.61]} \\
& \Exp 
& 1.22 {\color{blue}[1.18–1.25]} & 2.76 & 0.39 {\color{blue}[0.36–0.43]}
& 1.14 {\color{blue}[1.07–1.23]} & 2.47 & 0.46 {\color{blue}[0.36–0.51]} \\
& \Exp\ + \PT 
& 1.21 {\color{blue}[1.17–1.24]} & 2.71 & 0.40 {\color{blue}[0.37–0.44]}
& 1.12 {\color{blue}[1.09–1.20]} & 2.42 & 0.47 {\color{blue}[0.37–0.51]} \\
& \Ind 
& 1.24 {\color{blue}[1.19–1.28]} & 2.76 & 0.39 {\color{blue}[0.35–0.42]}
& 1.33 {\color{blue}[1.26–1.43]} & 3.03 & 0.33 {\color{blue}[0.24–0.36]} \\
& \Ind\ + \PT 
& 1.23 {\color{blue}[1.18–1.27]} & 2.72 & 0.40 {\color{blue}[0.36–0.43]}
& 1.33 {\color{blue}[1.28–1.41]} & 3.03 & 0.33 {\color{blue}[0.25–0.36]} \\
\bottomrule
\end{tabular}
\end{adjustbox}
\caption{Gemini 2.5 Pro results. 95\% confidence intervals (CIs, shown in blue) are reported.}
\label{apptab:cis_gemini}
\end{table*}

\section{Potential Pretraining Exposure}
\label{app:llmconta}

We observed that both \verb|Mixtral-8x7B| and \verb|Gemini 2.5 Pro| achieve notably low mean absolute errors (MAE $\approx$ 1.3 and 1.03 years, respectively) on age prediction over CHILDES transcripts.
While such precision may indicate genuine linguistic generalization, we cannot rule out the possibility that it is partially influenced by prior exposure to overlapping CHILDES--style transcripts during pretraining. To probe this possibility, we qualitatively examined a small set of responses (Table~\ref{apptab:exposure}) where models hypothesized specific corpora from contextual cues or produced meta-level statements suggesting familiarity with the CHILDES ecosystem. Importantly, these examples are anecdotal and do not constitute evidence of verbatim memorization, and self-reported exposure is not independently verifiable. Because pretraining data for these models are not fully disclosed, we interpret the results as consistent with potential dataset exposure rather than confirmed contamination.

\clearpage
\onecolumn

\begin{tcolorbox}[
  breakable,
  enhanced,
  colback=white,
  colframe=black!60,
  boxrule=1pt,
  arc=2pt,
  left=6pt,
  right=6pt,
  top=6pt,
  bottom=6pt,
  width=\textwidth
]
\small\ttfamily
Each adult utterance is marked as \texttt{[X-a]}, where \texttt{X} is a number indicating the turn order. Your task is to classify each of these adult utterances based on the categories below.

\medskip

\medskip
\textbf{Important:} You must determine whether an utterance is a question or not based solely on the content and structure, not on punctuation.

\medskip
First, determine whether the adult's utterance is a \textbf{direct question}.

\medskip
\hrulefill
\medskip

\textbf{If the utterance is a question, classify it as one of the following Question Types:}

\smallskip
\begin{itemize}
  \item \textbf{Display Question}: The adult already knows the answer and expects a specific factual response.
  \textit{(Does not include yes/no or choice questions)} \textit{e.g., ``What color is this apple?''}
  \item \textbf{Yes/No Question}: Questions explicitly requiring only a yes or no response.
  \textit{e.g., ``Did you go outside today?''}
  \item \textbf{Choice Question}: Questions explicitly asking the child to select one option from two or more provided choices.
  \textit{e.g., ``Do you want juice or milk?''}
  \item \textbf{Referential Question}: Questions genuinely seeking information about the child's personal experiences, opinions, or thoughts,
  with no predetermined correct answer.
  \textit{e.g., ``What did you do at school today?''}
  \item \textbf{Clarification Request}: Adult requests the child to repeat or clarify a previous utterance due to misunderstanding or inaudibility.
  \textit{e.g., ``What did you say?'', ``Huh?'', ``Can you say that again clearly?''}
  \item \textbf{Elaboration Prompt}: Questions or statements explicitly prompting the child to expand or provide additional details, reasons,
  or explanations about something previously mentioned.
  \textit{e.g., ``Can you tell me more about that?'', ``Why did that happen?''}
  \item \textbf{Confirmation Check}: Adult explicitly confirms their understanding of the child's previous utterance, usually by rephrasing or summarizing the child's words.
  \textit{e.g., ``Did you mean the red one?'', ``You're saying you went to the park?''}
\end{itemize}

\medskip
\textbf{If the utterance is not a question, classify it as one of the following Non-Question Types:}

\smallskip
\begin{itemize}
  \item \textbf{Topic Introduction}: Adult introduces a new topic, shares personal experiences, or describes past events to provide context or invite engagement, without asking a question.
  \textit{e.g., ``Today I saw a big dog.'', ``When I was a kid, I visited the mountains every summer.''}
  \item \textbf{Imperative/Request}: Adult gives a command or request intended to prompt an action, not to obtain information.
  \textit{e.g., ``Read please.'', ``Put that here.''}
  \item \textbf{Feedback/Evaluation}: Adult provides positive or negative evaluation or feedback on the child's behavior or utterance.
  \textit{e.g., ``Good job!'', ``That's not quite right.''}
  \item \textbf{Self-Talk/Thinking Aloud}: Adult speaks to themselves or expresses thoughts aloud, not directly addressing the child.
  \textit{e.g., ``Hmm, I can't remember where I put that.''}
\end{itemize}

\medskip
\textbf{Special Cases}

\smallskip
\begin{itemize}
  \item \textbf{Ambiguous/Unclear}: When the adult's utterance is completely incomprehensible or unintelligible enough that question type cannot be determined at all.
  \textit{e.g., ``xxx'', completely garbled speech, or utterances with no meaningful words, such as only ``um'', ``uh'', ``hm''}
\end{itemize}

\smallskip
\textbf{Note:} ``xxx'' marks unintelligible speech in the original transcript.

\medskip
\hrulefill

\medskip
\textbf{Output Format}

\medskip
Classify each \texttt{[X-a]} adult utterance using the categories above.
When multiple types are present, choose the type that represents the main communicative intent of the utterance.

\medskip
\textbf{IMPORTANT:} Use ONLY the exact categories defined above. Do not create new categories or add explanations in parentheses. Do not include any explanation, commentary, or scoring. 

\medskip
Follow the format exactly as shown below:

\medskip
\textbf{Format:}
\begin{quote}\ttfamily
[1-a] - Question Type: Yes/No Question \newline
[2-a] - Non-Question Type: Topic Introduction \newline
[3-a] - Question Type: Display Question
\end{quote}

\medskip
\hrulefill

\medskip
Now, analyze the following conversation: \{DIALOGUE\}
\end{tcolorbox}

\begin{center}
\begin{minipage}{\textwidth}
\captionsetup{type=table,hypcap=false}
\captionof{table}{System prompt template for classifying Previous Adult Utterance Type.}
\label{apptab:classification_prompt}
\end{minipage}
\end{center}

\begin{table*}[ht]
  \centering

  \begin{tcolorbox}[
    colback=white,
    colframe=black!60,
    boxrule=1.0pt,
    arc=2pt,
    left=6pt,
    right=6pt,
    top=6pt,
    bottom=6pt,
    width=\textwidth
  ]
  \small\ttfamily
  You are given a transcript of only the CHILD's utterances from a conversation.
  This transcript was converted from spoken audio. Ignore any empty or whitespace-only lines.

  \medskip
  \textbf{Task:}\\
  Estimate the child's age in years and months only.

  \medskip
  \textbf{Output format:}\\
  ``X years Y months'' (no extra words).

  \medskip
  \textbf{Examples:}\\
  -- 4 years 4 months\\
  -- 7 years 0 months

  \medskip
  \textbf{Transcript:}\\
  \{DIALOGUE\}

  \medskip
  \textbf{Output:}
  \end{tcolorbox}

  \caption{System prompt template used for LLM-based age prediction from child utterances.}
  \label{apptab:llm-age-prompt}
\end{table*}
\begin{table*}[ht]
  \centering
  \small
  \setlength{\tabcolsep}{4pt}
  \begin{tabular}{p{2.5cm} p{13.5cm}}
    \toprule
    \textbf{Rubric stage} & \textbf{Representative markers / regex patterns} \\
    \midrule
    Causal &
      \texttt{because, so, so that, since, therefore \dots} \\
    \midrule
    Contrast &
      \texttt{but, however, although, though, whereas, on the other hand, by the way, anyway \dots } \\
    \midrule
    Initiative &
      \texttt{let's, wanna, shall we, how about, maybe we can, do you want to, wanna try \dots} \\
    \bottomrule
  \end{tabular}
  \caption{Discourse marker categories}
  \label{apptab:markers-to-rubric}
\end{table*}

\clearpage
\twocolumn

\clearpage
\onecolumn

\begin{tcolorbox}[
  breakable,
  enhanced,
  colback=white,
  colframe=black!60,
  boxrule=1pt,
  arc=2pt,
  left=6pt,
  right=6pt,
  top=6pt,
  bottom=6pt,
  width=\textwidth
]
\small\ttfamily
Each child utterance is labeled with a number in square brackets (e.g., \texttt{[1]}, \texttt{[2]}, etc.). 
Please assign Linguistic Independence and Expansion scores to each based on the criteria provided below.

\medskip

\medskip

\medskip

\textbf{Step 1: Response Content Type}
\begin{itemize}
  \setlength{\itemsep}{2pt}
  \setlength{\parskip}{0pt}
  \item \textbf{Hesitation Only Response}: The child's utterance consists solely of hesitation/filler tokens (e.g., ``um,'' ``uh,'' ``hm,'' ``er,'' ``ah'') with no propositional or acknowledgment content.
  \item \textbf{Content Response}: The child's utterance contains meaningful words or content.
\end{itemize}

\medskip
\textbf{Step 2: Scoring}

\smallskip
\begin{itemize}
  \setlength{\itemsep}{2pt}
  \setlength{\parskip}{0pt}
  \item \textbf{For Hesitation Only Responses:}
  \begin{itemize}
    \setlength{\itemsep}{1pt}
    \setlength{\parskip}{0pt}
    \item Independence: 0 (automatic)
    \item Expansion: 0 (automatic)
  \end{itemize}

  \item \textbf{For Content Responses:} Apply the scoring criteria below.
\end{itemize}

\medskip
\textbf{Evaluation Criteria (For Content Responses Only)}

\medskip
\textbf{Note:} When a child's utterance includes both hesitation and real words, exclude the hesitation when evaluating both Linguistic Independence and Expansion scores. 
e.g., "um I don't know" → Score based only on "I don't know"
Focus only on the meaningful content words for all scoring purposes.
\medskip

\hrulefill

\medskip

\textbf{1) Independence (scoring criteria; 1 = very dependent, 10 = fully independent)}

\begin{itemize}
  \setlength{\itemsep}{2pt}
  \setlength{\parskip}{0pt}

  \item \textbf{Score 1 - Imitation without understanding (complete repetition)}: Directly repeats the adult's words identically; not a response. 
  \textit{e.g., Adult: Did you eat? → Child: Did you eat?} Key features: verbatim echo, no response content.

  \item \textbf{Score 2 - Partial imitation}: Repeats only the tail of the adult's utterance; does not supply content. 
  \textit{e.g., Adult: Did you eat red apple? → Child: Red apple?} Key features: echoes final words, no full comprehension.

  \item \textbf{Score 3 - Single acknowledgment only (no answer)}: Acknowledgment tokens only; not an answer to the question. 
  \textit{e.g., Yes, No, Okay, Uh-huh} Key features: acknowledgment only, no meaningful information.

  \item \textbf{Score 4 - Direct answer only (no initiative)}: Accurately answers what was asked, adds nothing else (no follow-up/evaluation/new info). 
  \textit{e.g., Red., At school., Two.} Key features: answer present, no initiative.

  \item \textbf{Score 5 - Answer + minimal initiative}: After answering, adds minimal non-exploratory move: closed follow-up, confirmation, clarification, or prompted reason (only after adult asks why). 
  \textit{e.g., Red. Which one?, At school--Is that okay?, What?, Because I like it. (after prompt)} Key features: minimal initiative, not open-ended, prompted reason allowed.

  \item \textbf{Score 6 - Self-initiated open-ended inquiry}: Independently advances with open-ended exploration (why/how/what-if/compare), with or without an answer. 
  \textit{e.g., Red--why is it red?, How does it work?, What if we pick blue?} Key features: self-driven inquiry, open-ended.

  \item \textbf{Score 7 - Spontaneous explanation (unprompted justification)}: Gives reasons/justifications without being asked why. 
  \textit{e.g., I picked red because it's faster.} Key features: unprompted justification.

  \item \textbf{Score 8 - Previous conversation connection}: Links prior talk/events to the current context with temporal continuity. 
  \textit{e.g., The friend I told you about--he's back!} Key features: across-turn linking, time continuity.

  \item \textbf{Score 9 - Topic management + listener-oriented regulation ($\ge 2$ features)}: In one turn performs $\ge 2$: topic open/shift, listener check, self-monitor/repair, close/return to prior topic. 
  \textit{e.g., Anyway, let's talk about the science project now. Does that make sense? I'll say it more slowly.} Key features: two or more control acts, functional regulation.

  \item \textbf{Score 10 - Pragmatic leadership with perspective-taking (integrated)}: Leads (proposes/organizes) and adapts to listener needs/state by integrating $\ge 2$: plan/proposal/role, perspective-taking, adaptive delivery. 
  \textit{e.g., Let's start with the Lego one; if you're tired, we can stop later. I'll explain slowly so you can follow.} Key features: plan plus listener adaptation, interactional leadership.
\end{itemize}

\textbf{2) Expansion (scoring criteria; 1 = no expansion, 10 = highly elaborated)}

\smallskip
\begin{itemize}
  \setlength{\itemsep}{2pt}
  \setlength{\parskip}{0pt}

  \item \textbf{Score 1 - Single word}: One noun/verb/adjective only. 
  \textit{e.g., Dog, Pretty, Dislike} Key features: word count=1.

  \item \textbf{Score 2 - Two words}: Modifier+noun or two separate words. 
  \textit{e.g., Cute dog, No that} Key features: word count=2.

  \item \textbf{Score 3 - 3–4 words, no sentence}: Word combination without full subject–verb structure. 
  \textit{e.g., Red pretty apple, Me like this} Key features: 3–4 words, no S–V structure.

  \item \textbf{Score 4 - Basic sentence}: One complete clause; no additional information. 
  \textit{e.g., The dog runs, I ate rice} Key features: single clause.

  \item \textbf{Score 5 - Sentence + modifiers}: Complete clause with 1–2 descriptive words. 
  \textit{e.g., The big dog runs fast} Key features: descriptive modifiers present.

  \item \textbf{Score 6 - Simple coordination (no causality)}: Uses and/but to coordinate clauses; no causal/conditional meaning. 
  \textit{e.g., I ate rice and played, It was fun but tiring} Key features: coordination only, no causality.

  \item \textbf{Score 7 - Single cause–effect}: One explicit causal link (because/so/therefore). 
  \textit{e.g., It was raining so I stayed inside} Key features: exactly one causal link.

  \item \textbf{Score 8 - Detailed sequence}: Chronological sequence of $\ge 3$ actions with ordering markers (then/after/first/next). 
  \textit{e.g., I went to the park, played on swings, then ate snacks} Key features: 3+ ordered events.

  \item \textbf{Score 9 - Complex logical structure}: Includes subordinating logic (if/when/unless/while/than) and remains coherent. 
  \textit{e.g., If it rains like today, I'll use an umbrella; if sunny, I'll walk} Key features: subordination present.

  \item \textbf{Score 10 - Multi-layered narrative}: Coherent linkage across multiple timeframes/perspectives or abstract connections. 
  \textit{e.g., Yesterday's rain reminded me of last summer when we couldn't go camping, which was good because we stayed home and I learned to bake} Key features: multiple time/perspective links.
\end{itemize}

\medskip
\textbf{Note:} "xxx" marks unintelligible speech in the original transcript.

\medskip
\hrulefill

\medskip
\textbf{Output Format}

\smallskip
For each child's utterance, provide scores in the following format.
Do not include any explanation, or commentary. Follow the format exactly as shown below:

\medskip
\textbf{For Content Responses:}

\smallskip
\texttt{[1] - Independence: X, Expansion: Y}

\texttt{[2] - Independence: X, Expansion: Y}

\medskip
\textbf{For Hesitation Only:}

\smallskip
\texttt{[1] - Independence: 0, Expansion: 0 [Hesitation Only]}

\texttt{[2] - Independence: 0, Expansion: 0 [Hesitation Only]}

\medskip
\textbf{Example Outputs:}

\smallskip
\texttt{[1] - Independence: 6, Expansion: 5}

\texttt{[2] - Independence: 0, Expansion: 0 [Hesitation Only]}

\texttt{[3] - Independence: 3, Expansion: 2}

\medskip
\hrulefill

\medskip
Now, analyze the following conversation: \{DIALOGUE\}
\end{tcolorbox}

\begin{center}
\begin{minipage}{\textwidth}
\captionsetup{type=table,hypcap=false}
\captionof{table}{System prompt template for scoring child responses in child--adult conversational transcripts, defining Linguistic Independence and Expansion.}
\label{apptab:scoring_prompt}
\end{minipage}
\end{center}

\clearpage
\twocolumn


\begin{table*}[t]
\centering
\small
\renewcommand{\arraystretch}{1.2}
\setlength{\tabcolsep}{5pt}
\begin{tabular}{p{3cm} p{11cm}}
\toprule
\textbf{Case} & \textbf{Transcript Excerpt and Model Responses} \\
\midrule

\textbf{(a) HSLLD corpus} &
\textit{Child: hey. there’s a penny and another penny...} \\
& \textit{Adult: it’s a magnet. see.} \\[2pt]
& \textbf{Mixtral:} \hl{``Corpus: HSLLD. Multiple speakers (Child, Mother) and object exploration indicate a lab-based design.''} \\
& $\rightarrow$ Infers the corpus from speaker composition and object-focused interaction, reflecting context-based reasoning. \\[2pt]
& \textbf{Gemini:} \hl{``This resembles a structured lab play session with a mother and investigator.''} \\
& $\rightarrow$ Identifies a lab-like play context and speaker roles, suggesting familiarity with similar transcript structures. \\
\midrule

\textbf{(b) Brown corpus} &
\textit{Adult: do you want some milk?} \\
& \textit{Child: want milk.} \\
& \textit{Adult: you can have it.} \\
& \textit{Child: mine.} \\[2pt]
& \textbf{Mixtral:} \hl{``Likely Brown corpus--home setting, short child utterances, and caregiver scaffolding.''} \\
& $\rightarrow$ Attributes the excerpt to Brown based on a home-style exchange, short child turns, and caregiver scaffolding. \\[2pt]
& \textbf{Gemini:} \hl{``This looks like a caregiver–child exchange during early development, possibly Brown, but uncertain.''} \\
& $\rightarrow$ Notes early developmental speech patterns and caregiver–child context while explicitly expressing uncertainty. \\
\midrule

\textbf{(c) Bliss corpus} &
\textit{Adult: your turn!} \\
& \textit{Child: let’s move this piece here.} \\
& \textit{Adult: good idea.} \\
& \textit{Child: I win!} \\[2pt]
& \textbf{Mixtral:} \hl{``Probably Bliss corpus--play-based discourse with initiative markers (‘let’s’, ‘your turn’).''} \\
& $\rightarrow$ Links the excerpt to a play-based interaction using initiative markers such as ``let’s'' and ``your turn.'' \\[2pt]
& \textbf{Gemini:} \hl{``This resembles CHILDES play-based dialogue involving cooperative games.''} \\
& $\rightarrow$ Characterizes the dialogue as play-based and cooperative, aligning it with a common CHILDES-style scenario. \\
\midrule

\textbf{(d) Awareness} &
\textit{Adult: do I get to play farm with you?} \\
& \textit{Child: yeah. they’re firefighters...} \\[2pt]
& \textbf{Mixtral:} \hl{``I cannot confirm the corpus but it resembles CHILDES toy-play transcripts (Brown or Bliss).''} \\
& $\rightarrow$ Avoids committing to a specific corpus while pointing to toy-play transcript structure typical of CHILDES. \\[2pt]
& \textbf{Gemini:} \hl{``Consistent with CHILDES toy-play transcripts, but corpus unknown.''} \\
& $\rightarrow$ Similarly recognizes CHILDES-like toy-play structure and acknowledges that the corpus cannot be determined. \\
\midrule

\textbf{(e) Self-Disclosure} &
\textit{Child: no no no. they eat chicken wings. and this they’re gonna have little ladybugs.} \\[2pt]
& \textbf{Mixtral:} \hl{(N/A--model did not mention corpus exposure.)} \\[2pt]
& \textbf{Gemini:} \hl{``As a large language model, I don’t ‘remember’ data, but the contents of the Ellis Weismer corpus, along with many other CHILDES corpora, were part of my pretraining dataset.''} \\
& $\rightarrow$ Self-reports possible exposure to CHILDES corpora during pretraining (not independently verifiable). \\
\bottomrule
\end{tabular}
\caption{Qualitative responses from \textbf{Mixtral-8×7B-Instruct} and \textbf{Gemini 2.5 Pro} that suggest potential pretraining exposure. Models sometimes hypothesize corpus identity from contextual cues or produce meta-level comments about CHILDES.}
\label{apptab:exposure}
\end{table*}

\end{document}